\documentclass[letterpaper, 10 pt, conference]{ieeeconf}

\IEEEoverridecommandlockouts
\usepackage{bm}
\usepackage{tikz}
\usepackage{times}
\usepackage{xcolor}
\usepackage{siunitx}
\usepackage{amssymb}
\usepackage{amsmath}
\usepackage{wrapfig}
\usepackage{flushend}
\usepackage{graphics}
\usepackage{hyperref}
\usepackage{multicol}
\usepackage{verbatim}
\usepackage{graphicx}
\usepackage{multirow}
\usepackage{booktabs}
\usepackage{pgfplots}
\usepackage{algorithm}
\usepackage{glossaries}
\usepackage{algorithmic}
\usepackage[caption=false, font=footnotesize]{subfig}
\usetikzlibrary{calc}
\usetikzlibrary{arrows.meta}
\usetikzlibrary{backgrounds}

\usepgfplotslibrary{groupplots}
\usepgfplotslibrary{fillbetween}
\pgfplotsset{compat=1.4}
\graphicspath{{fig/}}
\setacronymstyle{long-short}
\glsdisablehyper 
\DeclareMathAlphabet{\mathpzc}{OT1}{pzc}{m}{it}
\DeclareMathOperator{\SEDIST}{\Delta_\text{SE3}}
\DeclareMathOperator{\QUATDIST}{\Delta_\text{quat}}

\newcommand{\mdiff}[1]{#1}

\begin{document}

\title{\LARGE \bf
Uncertainty-Aware Self-Supervised Learning of \\
Spatial Perception Tasks
}

\author{Mirko Nava$^{1}$, Antonio Paolillo$^{1}$, J\'er\^ome Guzzi$^{1}$, Luca Maria Gambardella$^{1}$, and Alessandro Giusti$^{1}$ %
\thanks{This work was supported by the Swiss National Science Foundation (SNSF) through the NCCR Robotics.} %
\thanks{$^{1}$All authors are with the Dalle Molle Institute for Artificial Intelligence (IDSIA), USI-SUPSI, Lugano, Switzerland, {\tt\footnotesize mirko@idsia.ch}}%
}

\newacronym{ssl}{SSL}{self-supervised learning}
\newacronym[longplural={neural networks}, shortplural={NNs}]{nn}{NN}{neural network}
\newacronym[longplural={convolutional neural networks}, shortplural={CNNs}]{cnn}{CNN}{convolutional neural network}
\newacronym[longplural={recurrent convolutional neural networks}, shortplural={RCNNs}]{rcnn}{RCNN}{recurrent convolutional neural network}
\newacronym{ooi}{OOI}{object of interest}
\newacronym{icp}{ICP}{iterative closest point}

\maketitle


\begin{abstract}
We propose a general self-supervised \mdiff{learning approach for} spatial perception tasks, such as estimating the pose of an object relative to the robot, from onboard sensor readings.
The model is learned from training episodes, by relying on: a continuous state estimate, possibly inaccurate and affected by odometry drift; and a detector, that sporadically provides supervision about the target pose.  We demonstrate the general approach in three different concrete scenarios: a simulated robot arm that visually estimates the pose of an object of interest; a small differential drive robot using 7 infrared sensors to localize a nearby wall; an omnidirectional mobile robot that localizes itself in an environment from camera images. Quantitative results show that the approach works well in all three scenarios, and that explicitly accounting for uncertainty yields statistically significant performance improvements.
\mdiff{Videos, datasets, and code to reproduce our results are available at:
\url{https://github.com/idsia-robotics/uncertainty-aware-ssl-spatial-perception}.}
\end{abstract}


\section{Introduction}\label{sec:introduction}

Many robot perception tasks consist of interpreting sensor readings to extract high-level spatial information~\cite{jing2020tpam}, such as the pose of an \gls{ooi} with respect to the robot, or the pose of the robot itself in the environment.
When sensors produce noisy, high-dimensional data that is difficult to interpret (e.g. cameras or lidars), \mdiff{a common solution is to rely on supervised learning~\cite{song2020learning}}.
\mdiff{In many real-world scenarios, collecting the necessary training sets is a fundamental problem; \Gls{ssl} in robotics aims at equipping robots with the ability to acquire their own training data, e.g. by using additional sensors as a source of supervision, without any external assistance. In some cases, this allows robots to acquire training data directly in the deployment environment.}

\mdiff{In this context, a state estimator, such as a robot's odometry, can be a rich source of information.}
Odometry allows a robot to estimate its own motion in the environment according to its kinematics (e.g. by integrating over time the motion of its wheels as measured by wheel encoders), often with some uncertainty and accumulating errors. Consider for example a robot capable of odometry, equipped with a camera and a collision detector~\cite{gandhi2017learning}; after bumping into an object, the robot could reconsider the camera observations from the timesteps preceding the collision; assuming a static obstacle, the camera image acquired when the robot was (according to its odometry) \SI{1}{m} behind the place of collision, would depict an obstacle at a distance of \SI{1}{m}.  This piece of information was acquired by the robot without any explicit external supervision, and can be used for training machine learning models that map acquired images to the spatial position of obstacles.   The role of odometry is to leverage sparse information from a simple detector -- which provides relevant information only for a few timesteps in a training sequence -- to generate an informative labeled training set.

In this paper we generalize and extend this basic idea: we consider a generic robot that has a spatial perception task (e.g., estimating the pose of an \gls{ooi} in the environment), is capable of state estimation, possibly affected by accumulating uncertainty due to errors (e.g. odometry), and is equipped with one or more sensors, whose outputs we want to use to estimate the target pose.
Furthermore, the robot is equipped with a detector that, for at least a small fraction of timesteps, produces ground-truth information about the target pose (possibly uncertain).
Given training sequences, we want to learn a stateless model that, given the sensor readings, estimates the target pose.

\mdiff{%
After reviewing the related work in Sec.~\ref{sec:rw},
we illustrate our main contribution in  
Sec.~\ref{sec:model}: a formalization of this problem and a general solution based on deep learning, that: 
i) learns from sporadic supervision given by a detector and a possibly uncertain state estimator;
ii) can explicitly account for uncertainty in the state estimates and in the supervision using a Monte Carlo approach;
iii) integrates a recently proposed state-consistency loss~\cite{nava2021ral} to further improve results, even with the hurdle of uncertainty.
In Sec.~\ref{sec:exp_setup} we investigate the generality of our contribution by instantiating it in three different applications (more details in Table \ref{tab:envs}):

\begin{itemize}
\item Estimating the relative 3D pose of an \gls{ooi} from a camera mounted on a robotic arm manipulator.
\item Estimating the heading of a differential drive robot in the vicinity of a straight wall, using data from 7 sensors that measure the amount of infrared light reflected from the environment.
\item Estimating the 2D pose of a docking station, using images from a camera mounted on a mecanum robot.
\end{itemize}
Sec.~\ref{sec:results} experimentally evaluates the approach on the three applications and quantifies the improvements due to explicitly modeling uncertainty and enforcing state-consistency; while conclusions are drawn in Sec.~\ref{sec:conclusions}.
}

\begin{table*}
    \centering
    \caption{\mdiff{Platforms and sensory equipment used in the perception tasks}}
    \begin{tabular}{l@{}c@{}c@{}c@{}}
    \toprule
     \textbf{Robot} & Simulated Robot Manipulator & Real Differential Drive Robot & Real Mecanum Robot\\
     & %
     \begin{tikzpicture}[font=\normalsize, line width=1pt, text=white, color=white, style=every scope, scale=0.56] 
        \node (base) at (-2.5,-0.5){}; 
        \node (ee) at (-0.96,1.3){}; 
        \node (frame) at (3.7,0.9) {$\bm{x}(t)$};
        \node (ooi) at (-1.4,-2.2) {};
        
        \coordinate (up) at (0.0,0.8);
        \coordinate (right) at (0.65,0.38);
        \coordinate (forward) at (0.73,-0.3);
        
        \coordinate (ee_up) at (0.01,-0.7);
        \coordinate (ee_right) at (-0.65,-0.0);
        \coordinate (ee_forward) at (-0.05,0.34);
        
        \coordinate (ooi_right) at (-0.5,-0.5);
        \coordinate (ooi_up) at (-0.01,0.6);
        \coordinate (ooi_forward) at (0.6,-0.4);
        
        \begin{scope}[on background layer]
        \begin{scope}[style={line width=1pt}]
            \makeatletter\tikz@options\makeatother
            
            \node[inner sep=0pt] (image) at (0,0) {\includegraphics[height=0.5\columnwidth]{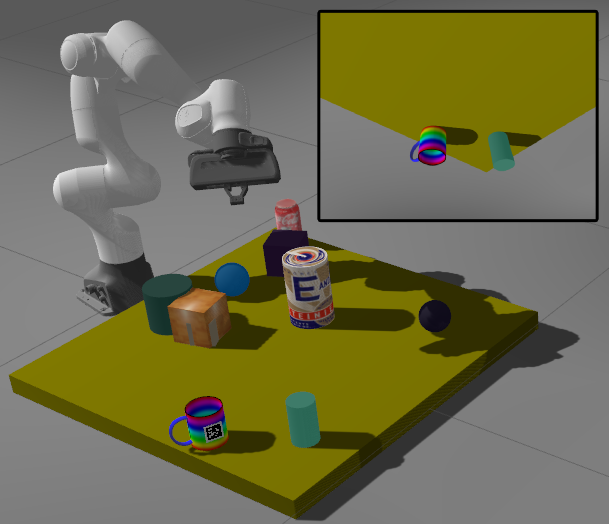}};
        
            \draw[-latex,color=blue] (base.center) -- node [left, near start] {{\color{white}${\cal F}_0$}} ++(up);
            \draw[-latex,color=red] (base.center) -- ++(forward);
            \draw[-latex,color=green!80!black] (base.center) -- ++(right);
            
            \draw[-latex,color=blue] (ee.center) -- ++(ee_up);
            \draw[-latex,color=red] (ee.center) -- ++(ee_forward);
            \draw[-latex,color=green!80!black] (ee.center) -- node [above, pos=0.45] {{\color{white}${\cal F}_r$}} ++(ee_right);
            
            \draw[-latex,color=red] (ooi.center) -- ++(ooi_right);
            \draw[-latex,color=blue] (ooi.center) -- node [left, pos=0.85] {{\color{white}${\cal F}_i$}} ++(ooi_up);
            \draw[-latex,color=green!80!black] (ooi.center) -- ++(ooi_forward);
        \end{scope}
        \end{scope}
        
        \draw[-{To}] (base) to [bend left=20] node[left,near end]{$\bm{p}(t)$} (ee);
        
        \draw[-{To}] (ee) to [bend left=30] node[right, pos=0.85]{$\bm{y}(t)$} (ooi);
     \end{tikzpicture} \hfill %
     & %
     \begin{tikzpicture}[font=\normalsize, line width=1pt, text=black, color=white, style=every scope, scale=0.66] 
        \node (base) at (-0.85,0.4){}; 
        \node (ee) at (-2.6,-1.8){}; 
        \node (frame) at (2.75,0.65) {$\bm{x}(t)$};
        \coordinate (up) at (0.0,0.8);
        \coordinate (right) at (0.9,-0.2);
        \coordinate (forward) at (-0.4,-0.7);
        \coordinate (ee_up) at (-0.05,0.8);
        \coordinate (ee_right) at (0.25,0.9);
        \coordinate (ee_forward) at (0.8,-0);
        \begin{scope}[on background layer]%
        \begin{scope}[style={line width=1pt}]%
            \makeatletter\tikz@options\makeatother%
            %
            \node[inner sep=0pt] (image) at (0,0) {\includegraphics[trim={0 0.5cm 3cm 2.8cm},clip,height=0.5\columnwidth]{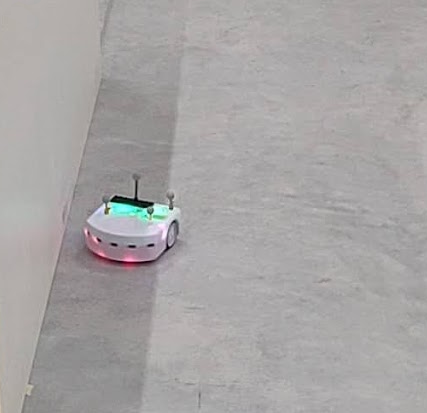}};%
            %
            \draw[-latex,color=blue] (base.center) -- node [left, pos=1.1] {{\color{white}${\cal F}_r$}} ++(up);%
            \draw[-latex,color=red] (base.center) -- ++(forward);%
            \draw[-latex,color=green!80!black] (base.center) -- ++(right);%
            %
            \draw[-latex,color=blue] (ee.center) -- node [right, pos=0.45] {{\color{white}${\cal F}_0$}} ++(ee_up);%
            \draw[-latex,color=red] (ee.center) -- ++(ee_forward);%
            \draw[-latex,color=green!80!black] (ee.center) --  ++(ee_right);%
        \end{scope}%
        \end{scope}%
        %
        \path [draw=black,line width=1pt,fill=white] (0.2,1.0) rectangle (3.255,3.0);%
        \node[text height=3mm] (sl) at (0.525,1.3){\color{black}\tiny\bf\texttt{L}};%
        \path [fill=blue] (0.345,1.4) rectangle (0.635,1.45);%
        \node[text height=3mm] (scl) at (0.925,1.3){\color{black}\tiny\bf\texttt{CL}};%
        \path [fill=blue] (0.745,1.4) rectangle (1.035,1.45);%
        \node[text height=3mm] (sc) at (1.375,1.3){\color{black}\tiny\bf\texttt{C}};%
        \path [fill=blue] (1.145,1.4) rectangle (1.435,1.8);%
        \node[text height=3mm] (scr) at (1.775,1.3){\color{black}\tiny\bf\texttt{CR}};%
        \path [fill=blue] (1.545,1.4) rectangle (1.835,2.6);%
        \node[text height=3mm] (sr) at (2.175,1.3){\color{black}\tiny\bf\texttt{R}};%
        \path [fill=blue] (1.945,1.4) rectangle (2.235,2.7);%
        \node[text height=3mm] (sbl) at (2.575,1.3){\color{black}\tiny\bf\texttt{BL}};%
        \path [fill=blue] (2.345,1.4) rectangle (2.635,1.45);%
        \node[text height=3mm] (sbr) at (2.975,1.3){\color{black}\tiny\bf\texttt{BR}};%
        \path [fill=blue] (2.845,1.4) rectangle (3.135,1.45);%
        %
        \draw[{To}-] (base) to [bend left=80] node[right,pos=0.5]{$\bm{p}(t)$} (ee);
        %
        \draw[{To}-] (ee) to [bend left=90] node[left, pos=0.6]{$\bm{y}(t)$} (base);
     \end{tikzpicture} \hspace{.4mm} %
     & %
      \begin{tikzpicture}[font=\normalsize, line width=1pt, text=black, color=white, style=every scope, scale=0.64] 
        \node (docking) at (-2.5,-2.5){}; 
        \node (robot) at (1.4,-1.2){};
        \node (frame) at (-3.75,1.2) {$\bm{x}(t)$};
        \coordinate (docking_z) at (0.0,0.8);
        \coordinate (docking_x) at (0.65,0.4);
        \coordinate (docking_y) at (-0.8,0.4);
        \coordinate (robot_z) at (0.01,0.7);
        \coordinate (robot_x) at (-0.65,0.5);
        \coordinate (robot_y) at (-0.65,-0.3);
        \begin{scope}[on background layer]%
        \begin{scope}[style={line width=1pt}]%
            \makeatletter\tikz@options\makeatother%
            %
            \node[inner sep=0pt] (image) at (0,-0.05) {\includegraphics[height=0.5\columnwidth]{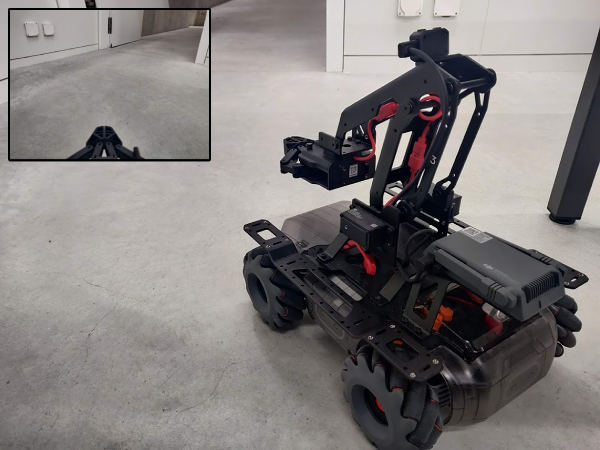}};%
            %
            \draw[-latex,color=blue] (robot.center) -- node [right, pos=-0.1] {{\color{white}${\cal F}_r$}} ++(robot_z);%
            \draw[-latex,color=red] (robot.center) -- ++(robot_x);%
            \draw[-latex,color=green!80!black] (robot.center) -- ++(robot_y);%
            %
            \draw[-latex,color=blue] (docking.center) -- node [left, pos=-0.2] {{\color{white}${\cal F}_0$}} ++(docking_z);%
            \draw[-latex,color=red] (docking.center) -- ++(docking_x);%
            \draw[-latex,color=green!80!black] (docking.center) --  ++(docking_y);%
        \end{scope}%
        \end{scope}%
        \draw[{To}-] (robot) to [bend left=50] node[above,pos=0.5]{$\bm{p}(t)$} (docking);%
        \draw[{To}-] (docking) to [bend left=50] node[below, pos=0.65]{$\bm{y}(t)$} (robot);%
     \end{tikzpicture}%
     \\ [2pt]%
     \textbf{Sensor} & RGB Camera & Infrared Sensors & RGB Camera \\ [2pt]
     \textbf{Target} & Object of Interest 3D Pose & Robot Heading & Docking Station 2D Pose \\ [2pt]
     \textbf{\begin{tabular}{@{}l}Detector\end{tabular}} & Fiducial Marker & Initial Robot Pose & Initial Robot Pose \\ [2pt]
     \textbf{\begin{tabular}{@{}l}State \\ [-2pt] Estimator\end{tabular}} & Forward Kinematics & Odometry (inaccurate) & Odometry (inaccurate) \\[3pt] 
     \textbf{Section} &
     \ref{subsec:exp_setup_panda}, \ref{subsec:results_panda} &
     \ref{subsec:exp_setup_wall}, \ref{subsec:results_wall} &
     \ref{subsec:exp_setup_station}, \ref{subsec:results_station} \\
    \bottomrule
    \end{tabular}
    \label{tab:envs}%
\end{table*}


\section{Related Work}\label{sec:rw}

\subsection{Self-Supervised Learning}

\Gls{ssl} refers to approaches that utilize a subset of the available features, or an elaboration of those, as supervisory information~\cite{jing2020tpam}.
In robotics, the term is also used to denote an autonomous, unattended robot that collects data, from which both input features and supervisory labels are extracted.
While many \gls{ssl} approaches focus on specific systems to extract the target variable~\cite{lieb2005adaptive, gandhi2017learning, kouris2018learning, pinto2016supersizing, jang2018grasp2vec, deng2020self, nava2019ral, wellhausen2019should, zurn2020self}, no explicit notion of a general detector is present.
In navigation, supervisory information is extracted from the knowledge that at time $t = 0$ a vehicle is on the road~\cite{lieb2005adaptive},
from letting a drone roam the environment until a crash happens~\cite{gandhi2017learning}, or by continuously measuring the distance from the surroundings~\cite{kouris2018learning}.
Similar to our approach, Lieb et al.~\cite{lieb2005adaptive} and Gandhi et al.~\cite{gandhi2017learning} reconstruct the ground-truth from a small subset of data, respectively at the beginning and at the end of each episode; 
while Kouris et al.~\cite{kouris2018learning} utilize a dense ground-truth estimate generated from three laser sensors.
In grasping, supervision is derived from measuring the force perceived on the end effector before and after a grasp attempt~\cite{pinto2016supersizing} or by iteratively refining the 3D pose prediction of known objects~\cite{deng2020self}.
Pinto et al.~\cite{pinto2016supersizing} predict the probability of grasping an object from 18 possible angles using images. Ground-truth information is generated once per each grasping attempt;
while Deng et al.~\cite{deng2020self} utilize a pre-trained \gls{cnn} to serve as a continuous source of ground-truth information.
The model learns online, and by grasping and moving objects produces more data, repeating the cycle.
In traversability estimation, a dense detector is derived from future or past sensors readings \cite{nava2019ral, wellhausen2019should, zurn2020self}.
Nava et al.~\cite{nava2019ral} learn a robot-centric obstacle map composed of traversable and obstacle cells. Labels are generated for each timestep considering proximity sensors' readings collected over a sliding time window.
\mdiff{Both sensors' readings and odometry are assumed to be ideal, while in our work we deal with uncertain (noisy) information.}
Wellhausen et al.~\cite{wellhausen2019should} predict terrain class and a traversability score by relating the future torque measured on the legs of the ANYmal quadruped, to image space footholds perceived currently.
Similarly, Z{\"u}rn et al.~\cite{zurn2020self} learn to classify the terrain from images, using as supervision the Fourier transform of sound captured near the wheels of a ground robot.
In the visual odometry field, self-supervised approaches extract ground-truth information directly from ego motion, learning the relative transformation between two images collected by a camera~\cite{agrawal2015learning, iyer2018geometric}, or from inverse warping images collected at time $t + 1$ and  $t - 1$ with respect to the current image~\cite{zhou2017unsupervised}.
Iyer et al.~\cite{iyer2018geometric} propose a geometric consistency term, aimed at improving the performance of the visual odometry module: they enforce the concatenation of relative motions from time $t$ to $t + n$ to be close to the motion of $t + n$ with respect to $t$.
In contrast, the state-consistency loss~\cite{nava2021ral} adopted in this work, is not limited to ego-motion information and enforces consistency on a general model's prediction 
\mdiff{(i.e., can be applied to pose estimation of the robot itself or other objects in the environment).
Additionally, \cite{iyer2018geometric} require the use of ad-hoc SE(3) Layers in the \gls{nn}, which represent the pose as a 3D rigid body transformation matrix, while the proposed work does not require any specific \gls{nn} architecture, nor a particular pose representation and associated distance function.}

\mdiff{%
\subsection{Noisy Data in Regression}

In real world scenarios, especially when data is collected directly by robots, noise present in the labels can severely degrade the performance of learned models \cite{song2020learning}.
Most approaches that do regression from noisy data focus on estimating the uncertainty associated with predictions.
These approaches fall under the category of Bayesian learning: some apply an uncertainty estimation model on an existing regressor \cite{loquercio2020general}, while others design ad-hoc architectures, casting the problem as learning the distribution of the model weights \cite{blundell2015weight, gal2016dropout, yeming2018flipout}.

Loquercio et al. \cite{loquercio2020general} estimate the uncertainty associated with model predictions by fitting a Bayesian belief network.
Dropout is used during multiple forward passes of the network to approximate the uncertainty of the prediction in a Monte Carlo sampling fashion.
Similarly, Gal et al. \cite{gal2016dropout} propose to approximate the true uncertainty by utilizing dropout Monte Carlo samples.
In contrast to \cite{loquercio2020general}, they require to alter the network architecture by placing dropout layers after each non-linearity.
By keeping the dropout functionality active during inference, they compute the uncertainty as the variance of the produced samples.
Blundell et al. \cite{blundell2015weight} propose to model \gls{nn} weights with a zero-centered Gaussian distribution.
Learning is then casted as a variational inference problem, where the true target distribution is approximated by the learned model conditioned on input data.
By noting that commonly used weight distributions are symmetric and independent, 
Yeming et al. \cite{yeming2018flipout} improve the ideas of \cite{blundell2015weight} by decomposing the forward pass into the multiplication of input and weight means by independently sampled sign matrices \cite[eq. (4)]{yeming2018flipout}, resulting in faster computation and less variance in the gradients.

In contrast, our approach tackles the different problem of learning from noisy data, without delving into the estimation of the uncertainty associated with predictions.
}

\subsection{Related Case Studies}

We demonstrate the generality of our approach by solving common tasks in the robotics field: \gls{ooi} pose estimation with a robotic arm, and localization of mobile ground robots.

Zeng et al.~\cite{zeng2017multi} use \gls{ssl} to collect RGB-D images of single objects from different points of view, then segmented by a background removal algorithm. Labels are then extracted by fitting the 3D model of the object in the corresponding image using an \gls{icp} algorithm.
Deng et al.~\cite{deng2020self} iteratively refine the performance of a \gls{nn} tasked to predict the pose of known objects from RGB data with online self-supervision: initially the network is trained on simulated data; it is then deployed in a real-world scenario where a camera-equipped robotic arm moves by pushing or grasping the objects around, generating new data used to improve the pose estimation.
Both approaches rely on 3D models of the objects, used during the learning process to refine the prediction, overlaying the 3D object on top of the image and measuring the difference.
Instead, our approach does not require prior knowledge of the object and relies solely on a detector capable of recognizing the object, or a fiducial marker attached to it.

Ratz et al.~\cite{ratz2020oneshot} localize the robot within an environment from LiDAR scans and camera frames.
A \gls{nn} fuses together the two sensors and produces a multi-modal descriptor. Localization is done in a one-shot fashion by matching the descriptor with a pre-computed database.
In our case studies, the approach utilizes only camera images and directly learns to localize, internally learning a discriminative feature space for the environment where data are collected.


\section{Model}\label{sec:model}

\subsection{Definitions}
We aim to train a model that, given readings $\bm{x}(t)$ collected by onboard sensors, predicts a target variable $\bm{y}(t)$, which is a pose in SE(3) of the frame ${\cal F}_i$ of an \gls{ooi}, relative to the moving robot frame ${\cal F}_r$. The sensor readings $\bm{x}(t)$ do not need to have an explicit geometric interpretation, might be high-dimensional (e.g. an uncalibrated image), and could potentially represent the concatenated outputs from multiple heterogeneous sensors.

To train the model, we use data collected in one or more training episodes. During each episode, the \gls{ooi} is static (i.e. $\bm{y}$ does not change when expressed in a fixed reference frame) while the robot moves in the environment. Let $\mathcal{T}$ be the set of all timesteps in a given training episode.

\begin{figure}
    \centering
    \includegraphics[width=\linewidth]{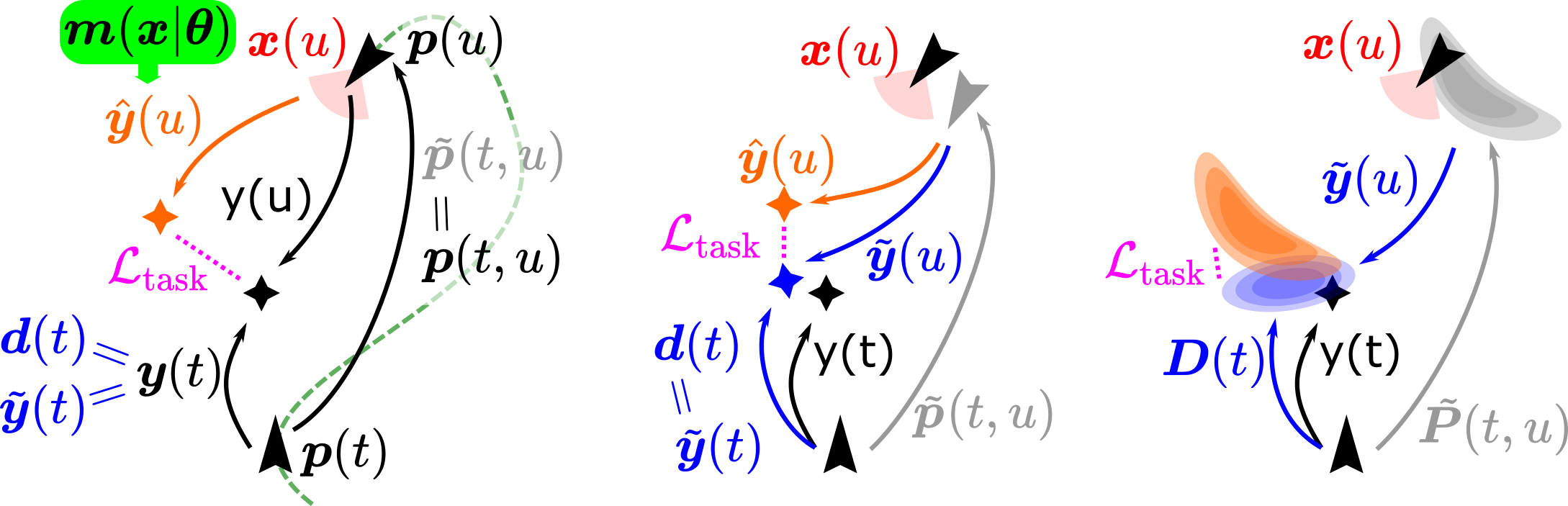}
    \caption{Illustration of the task loss in case the detector and odometry are ideal (left), inaccurate (center), \mdiff{or with a known uncertainty model (right).  Black color denotes the true poses for the robot (arrowhead) and \gls{ooi} (diamond); blue color denotes the \gls{ooi} pose as returned by the detector; orange denotes model predictions, obtained from data (red) sensed at time $u$.  Gray denotes robot odometry. See text for details.}}
    \label{fig:model1}
\end{figure}

We assume that a black-box \emph{detector} 
provides temporally sparse estimates of $\bm{y}(t)$:
\begin{equation}
\bm{d}(t) =
\begin{cases}
\Tilde{\bm{y}}(t) & \text{if } t \in \mathcal{T}_d, \\
\text{undefined } & \text{otherwise}
\end{cases}
\end{equation}
where $\mathcal{T}_d \subseteq \mathcal{T}$ denotes the set of timesteps in which the detector module provides an output, and the tilde over $\bm{y}$ denotes the fact that this is a potentially inaccurate estimate of the true value of the target variable.

Finally, we assume that an odometry module outputs for every $t \in \mathcal{T}$ a (potentially inaccurate) estimate $\Tilde{\bm{p}}(t)$ of the robot pose $\bm{p}(t)$ in a fixed inertial frame ${\cal F}_0$.  Given two timesteps $t, u \in \mathcal{T}$, we denote with $\Tilde{\bm{p}}(t,u)$ the odometry's estimate of the pose of the robot at $u$ with respect to its pose at $t$.  In particular, $\Tilde{\bm{p}}(t,u) = \ominus \bm{p}(t) \oplus \bm{p}(u)$, where $\oplus$ denotes the pose composition operator, and the unary operator $\ominus$ denotes pose inversion~\cite{corke2017robotics}.

For each training episode, we collect the set of samples
\begin{equation}
    \left\{\bm{x}(t), \bm{p}(t), \bm{d}(t) | t \in \mathcal{T}\right\};
\end{equation}
and use data from all training episodes to learn a mapping from $\bm{x}$ to $\bm{y}$.  
\mdiff{The mapping is implemented by a \gls{nn} model $\bm{m}(\bm{x} | \bm{\theta})$ parametrized by $\bm{\theta}$. \mdiff{Training is performed by minimizing a loss function}%
\begin{equation}\label{eq:scloss}
     \mathcal{L} = \mathcal{L}_\text{task} + \lambda_\text{sc} \mathcal{L}_\text{sc},
\end{equation}
composed by a task loss $\mathcal{L}_\text{task}$ and a state-consistency loss $\mathcal{L}_\text{sc}$; the latter is scaled with a factor $\lambda_\text{sc}$.}

\begin{figure}
    \centering
    \includegraphics[width=\linewidth]{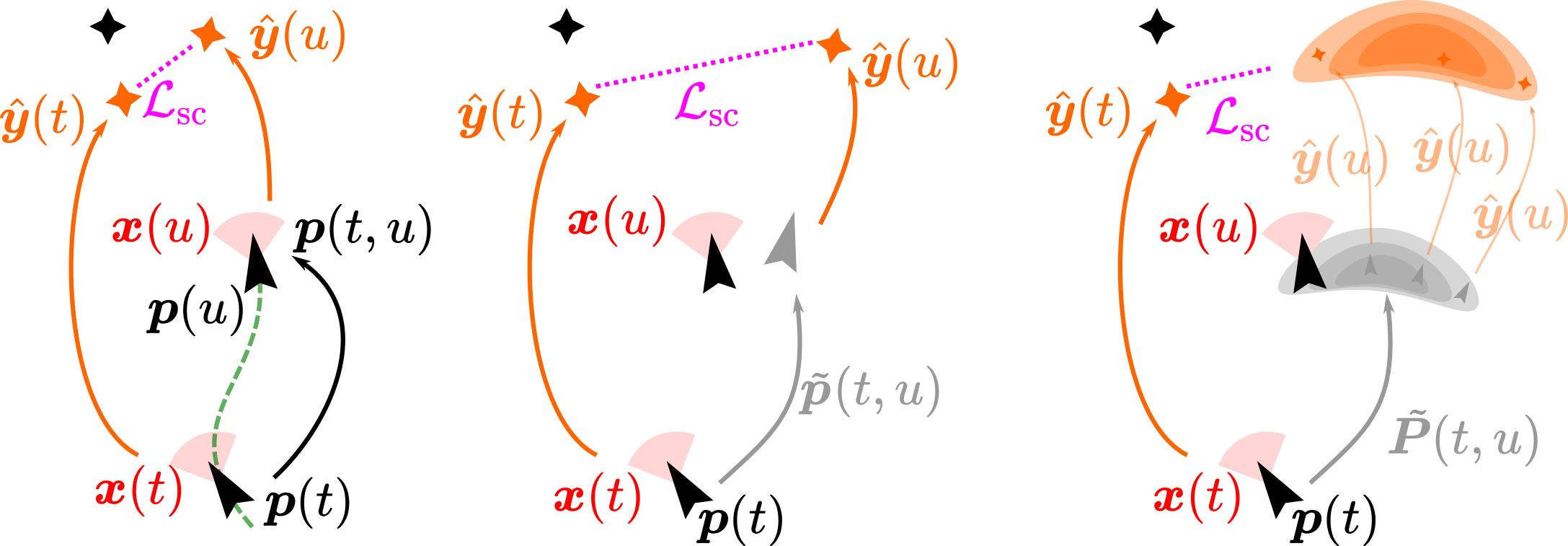}
    \caption{Illustration of the state-consistency loss in case the odometry is ideal (left), inaccurate (center), \mdiff{or with a known uncertainty model (right); black color denotes the true poses for the robot (arrowhead) and \gls{ooi} (diamond).  Orange denotes the outputs of the model at times $t$ and $u$, which depends on sensed data (red). The state consistency loss (violet) forces the model to output consistent estimates, accounting also for odometry (gray) and its uncertainty, if known.}}
    \label{fig:model2}
\end{figure}

We first describe these two terms in case the detector and odometry return point-wise estimates; then, we extend the discussion to the case in which their uncertainty can be modeled with a probability distribution, see Fig.~\ref{fig:model1} and Fig.~\ref{fig:model2}.

\subsection{Task loss}
Consider two timesteps $t,u$ in the same episode, such that $t \notin \mathcal{T}_d$ and $u \in \mathcal{T}_d$.
The task loss enforces that the model, when fed with $\bm{x}(t)$ returns a pose that is consistent with $\bm{d}(u)$,
accounting for the pose transform measured between $t$ and $u$ by the robot odometry, i.e. $\Tilde{\bm{p}}(t,u)$.

More specifically, $\bm{d}(u)$ is an estimate of the target variable with respect to the robot frame at time $u$; it follows that $\Tilde{\bm{y}}(t) = \Tilde{\bm{p}}(t,u) \oplus \bm{d}(u)$ is an estimate of the target variable at $t$. Thus, the task loss is defined as
\begin{equation}
\mathcal{L}_\text{task} = \sum_{\substack{t \in \mathcal{T}_d, u \in \mathcal{T}}} \SEDIST(\Tilde{\bm{p}}(t,u) \oplus \bm{d}(u), \bm{m}(\bm{x}(t)|\bm{\theta})),
\label{eq:task_loss}
\end{equation}
where the function $\SEDIST(\cdot,\cdot)$ is a measure of the distance between two poses in SE(3) and defined as
\begin{equation}\label{eq:se3dist}
    \SEDIST(\bm{p}_a,\bm{p}_b) := \lambda_\text{o} \left \Arrowvert \bm{o}_a - \bm{o}_b \right \Arrowvert + \frac{1}{\pi} \, \QUATDIST \left( \bm{q}_a, \bm{q}_b \right) , 
\end{equation}
where $\bm{p}_a$ and $\bm{p}_b$ are two generic poses, composed of the position components $\bm{o}_{a},\bm{o}_{b} \in \mathbb{R}^3$, and the rotation components represented as quaternions $\bm{q}_a, \bm{q}_b \in \mathbb{H}$, being $\mathbb{H}$ the non-commutative ring of the quaternions.
In~\eqref{eq:se3dist}, $\QUATDIST(\cdot, \cdot)$ denotes the quaternionic distance \cite[eq. (4)]{mahendran20173d}.
Note that the rotational term of the distance is bound in $[0, 1]$ while the positional term has no upper bound.  The parameter $\lambda_\text{o}$ is introduced as a scaling factor to weigh the two terms.\footnote{\mdiff{%
In principle, other options for representing poses and their distance \cite{zhou2019continuity, peretroukhin2020smooth} might be adopted, as long as the distance function is continuous and derivable.}}

This definition of the task loss uses odometry to propagate the estimate of $\bm{y}$ (produced by the detector in a timestep $u \in \mathcal{T}_d$), to any other timestep $t$ in the same episode.  If both the detector and the odometry are ideal (i.e. error-free), using any $u \in \mathcal{T}_d$ yields the same value of $\Tilde{\bm{y}}(t) = \Tilde{\bm{p}}(t,u) \oplus \bm{d}(u)$.  Otherwise, if the detector and/or the odometry are not ideal, every different choice of $u$ yields a different estimate of $\Tilde{\bm{y}}(t)$. In practice, this is expected to mitigate inaccuracies as errors are averaged out during training.

\subsection{State-Consistency Loss}\label{subsec:scloss}

Consider two timesteps $t,u$ in the same sequence, and assume that $t, u \notin \mathcal{T}_d$.  The state-consistency loss enforces that the predictions of the model at $t$ and $u$ are consistent with each other~\cite{nava2021ral}, accounting for the robot's odometry between $t$ and $u$. More specifically, 
\begin{equation}
    \mathcal{L}_\text{sc} = \sum_{\substack{t, u \in \mathcal{T}}} \SEDIST(\Tilde{\bm{p}}(t,u)\oplus \bm{m}(\bm{x}(u)|\bm{\theta}), \bm{m}(\bm{x}(t)|\bm{\theta})).
    \label{eq:loss_sc}
\end{equation}

Consider the following example; given $\bm{x}(t)$, the model returns an estimated pose for an \gls{ooi} \SI{1.5}{m} in front of the robot; after the robot advances \SI{1}{m}, at time $u$, the model given $\bm{x}(u)$ should return a pose that is \SI{0.5}{m} in front of the robot.
The state-consistency loss ensures that predictions $\hat{\bm{y}}(t) \! = \! \bm{m}(\bm{x}(t)|\bm{\theta})$ and $\hat{\bm{y}}(u) \! = \! \bm{m}(\bm{x}(u)|\bm{\theta)}$ match this expectation.

\subsection{Dealing with uncertainty}\label{subsec:uncertainty}

Our approach relies on two sources of information, namely the detector $\Tilde{\bm{d}}(t)$ and the odometry $\Tilde{\bm{p}}(t,u)$, both of which are possibly affected by measurement errors (instantaneous for the detector, accumulated over time for the odometry).
If such errors can be modeled, we can explicitly account for them in our approach.

In particular, we represent the uncertainty of $\bm{d}(t)$ by considering that the detector's output is, instead of a pointwise estimate of the target pose, a probability distribution over poses, defined in SE(3).  We denote such probability distribution as $\bm{D}(t)$.

Similarly, for odometry we define $\bm{P}(t,u)$ as the probability distribution of the relative pose of $\bm{p}(t,u)$; this also accounts for the fact that odometry errors accumulate over time, and therefore are not independent for different times. 

This representation allows to reformulate the task loss as
\begin{equation}
\begin{aligned}
\mathcal{L}_\text{task} = 
\sum_{\substack{t \in \mathcal{T}_d\\u \in \mathcal{T}}} 
\mathbb{E}\bigg[\SEDIST\Big(\bm{p}(t,u) \oplus \bm{d}(t), \bm{m}(\bm{x}(u)|\bm{\theta})\Big)\bigg],
\end{aligned}
\label{eq:task_loss_unc}
\end{equation}
and similarly, the state-consistency loss can be rewritten as
\begin{equation}
\begin{aligned}
\mathcal{L}_\text{sc} = 
\sum_{\substack{t, u \in \mathcal{T}}}
\mathbb{E}\bigg[\SEDIST\Big(\bm{p}(t,u) \oplus \bm{m}(\bm{x}(u)|\bm{\theta}), \bm{m}(\bm{x}(t)|\bm{\theta})\Big)\bigg]
\end{aligned}
\label{eq:loss_sc_unc}
\end{equation}
where $\bm{p}(t,u) \sim \bm{P}(t,u)$ and $\bm{d}(t) \sim \bm{D}(t)$.

In practice, when implementing the losses we approximate the expectation as the average over a finite number of realizations, in which $\bm{d}(t)$ and $\bm{p}(t,u)$ are Monte Carlo samples of the respective distributions.\footnote{\mdiff{In a straightforward implementation, this multiplies the number of training samples by a factor equal to $N_{\text{mc}}$, and yields a correspondingly longer training time;
in contrast, no additional computation is needed during inference.
}}


\section{Experimental setup}\label{sec:exp_setup}

This section validates our approach with three applications that differ in complexity and input dimensionality:
(i) pose estimation of an \gls{ooi} with a robotic arm; (ii) robot heading estimation using infrared sensors; and (iii) indoor localization of a ground robot. 

\subsection{Object of Interest Pose Estimation with a Robotic Arm} \label{subsec:exp_setup_panda}
We consider the robotic manipulator Panda by Franka Emika, equipped with an Intel RealSense D435 sensor~\cite{keselman2017intel} at its end-effector, simulated with Gazebo~\cite{koening2004degsign}, see Table~\ref{tab:envs}.
The task is to estimate the 3D pose of an \gls{ooi}, i.e. a colored mug, which is equipped with a small visual fiducial marker that is visible only when the cup is observed from a specific viewpoint.
The frames ${\cal F}_0$ and ${\cal F}_r$ are placed at the base of the robot and at its end-effector respectively; while the frame ${\cal F}_i$ of the \gls{ooi} is in the center of the mug.
The end-effector pose $\bm{p}(t)$ is given by the robot forward kinematics.
The input $\bm{x}(t)$ is a $160 \times 120$~pixel RGB image acquired from the RealSense camera.
The estimates $\bm{d}(t)$ are generated by the AprilTag~\cite{wang2016apriltag} off-the-shelf fiducial marker detector operating on $\bm{x}(t)$;  only when the marker is clearly visible in the frame, the detector returns a noisy estimate of the 3D pose of the marker w.r.t. ${\cal F}_r$\footnote{The rigid transformation between the end-effector and the camera frame is assumed to be known, e.g., through a calibration procedure.
Similarly, the offset between the origin of ${\cal F}_i$ and the center of the marker is taken into account in our computations.}. 

As shown in Table~\ref{tab:envs}, the environment consists of a flat table of 90$\times$90~{cm} with different objects, some textured and some others having a solid color, besides the mug. 
In each training episode, the table and objects color, the position of the objects, and the direction of the light illuminating the scene are randomized to generate different environments.
For each environment, the robot moves the end-effector in order to reach a total of 32 goal poses using the ROS MoveIt \cite{coleman2014reducing} implementation of the RRT planner.
Each goal position is sampled from a semi-sphere having a radius of \SI{55}{cm} placed at a height of \SI{35}{cm} from the table; the goal orientation is set to make the camera look towards a random point lying \SI{5}{cm} above the table.
For each environment, the end-effector pose is initialized at the center of the semi-sphere.

The collected data amounts to 237k tuples (of which only 78k have the mug visible), corresponding to 157 environments and 5 simulated hours.
The data is finally split into a training set (119 environments), a validation set (18 environments), and a testing set (20 environments).
Training is performed with a $\lambda_\text{o} = 10$, striking a balance between the positional and rotational errors.

\subsection{Robot Heading Estimation using Infrared Sensors}\label{subsec:exp_setup_wall}
  
For this application, we use a Thymio~\cite{mondada2017ram}, a small differential drive robot equipped with $7$ infrared sensors: $5$ mounted at the front and $2$ at the rear of the robot body. Each sensor measures the amount of infrared light reflected from the environment, which is related in some unknown way to the distance and orientation of the sensor with respect to an obstacle.
The input of the model $\bm{x}(t)$ consists in the uncalibrated readings of the 7 sensors at time $t$, while $\bm{y}(t)$ is the angle of the wall w.r.t the robot.
Note that we can still adopt $\SEDIST$, setting $\lambda_\text{o} = 0$, thus considering only the heading.
The robot odometry $\Tilde{\bm{p}}(t)$, derived from the wheel encoders, provides the 2D transformation of the robot frame ${\cal F}_r$ w.r.t. the inertial frame ${\cal F}_0$.
The scenario, along with an illustrative schematic of the sensor readings, is shown in Table~\ref{tab:envs}. 

Episodes are collected by teleoperating the robot along trajectories in the proximity of the wall.  During each episode, the robot true pose is tracked by a fixed tracking infrastructure (12 Optitrack cameras), which is used as a comparison for experiments.
At the beginning of each episode, the robot touches the wall with its rear side, thus the inertial frame ${\cal F}_0$ coincides with the robot frame ${\cal F}_r$ at this instant in time.
This piece of information also acts as a virtual detector, whose output $\Tilde{\bm{d}}(t)$ is available only in the first timestep of each episode.

Information about the rotation of each wheel is computed by the robot's firmware by measuring the current flowing through each motor -- an inaccurate approach whose errors we model to compute the robot's uncertain odometry.

A total of $16$ distinct episodes are recorded, each lasting on average $34$ seconds, during which samples are collected at $10$ Hz (a total of 5453 samples).  Episodes are split into training and validation sets for experiments using a leave-one-episode-out cross-validation scheme.

\subsection{Indoor Localization of a Ground Robot}\label{subsec:exp_setup_station}

We consider a wheeled ground robot, the DJI RoboMaster EP, equipped with an onboard camera and omnidirectional motion capabilities using Swedish wheels, see Table~\ref{tab:envs}.
We consider a situation in which the robot navigates a given indoor environment and, when required, needs to come back to a fixed docking station (e.g. to recharge the batteries).
In our setting, the docking station is represented by a mark on the floor.
The input $\bm{x}(t)$ is the camera stream, downsampled to a resolution of $160\times 120$~pixels; the perception task consists in predicting the pose of the docking station relative to the robot frame ${\cal F}_r$, given an image $\bm{x}(t)$ acquired at a generic pose.
Note that, since the docking station is fixed in the environment, this problem is equivalent to robot localization; it differs from the object localization task presented in~Sec.\ref{subsec:exp_setup_panda} because the docking station does not need to be visible in the input image for successful estimation.

\begin{figure}
    \centering
    \includegraphics[width=0.82\linewidth]{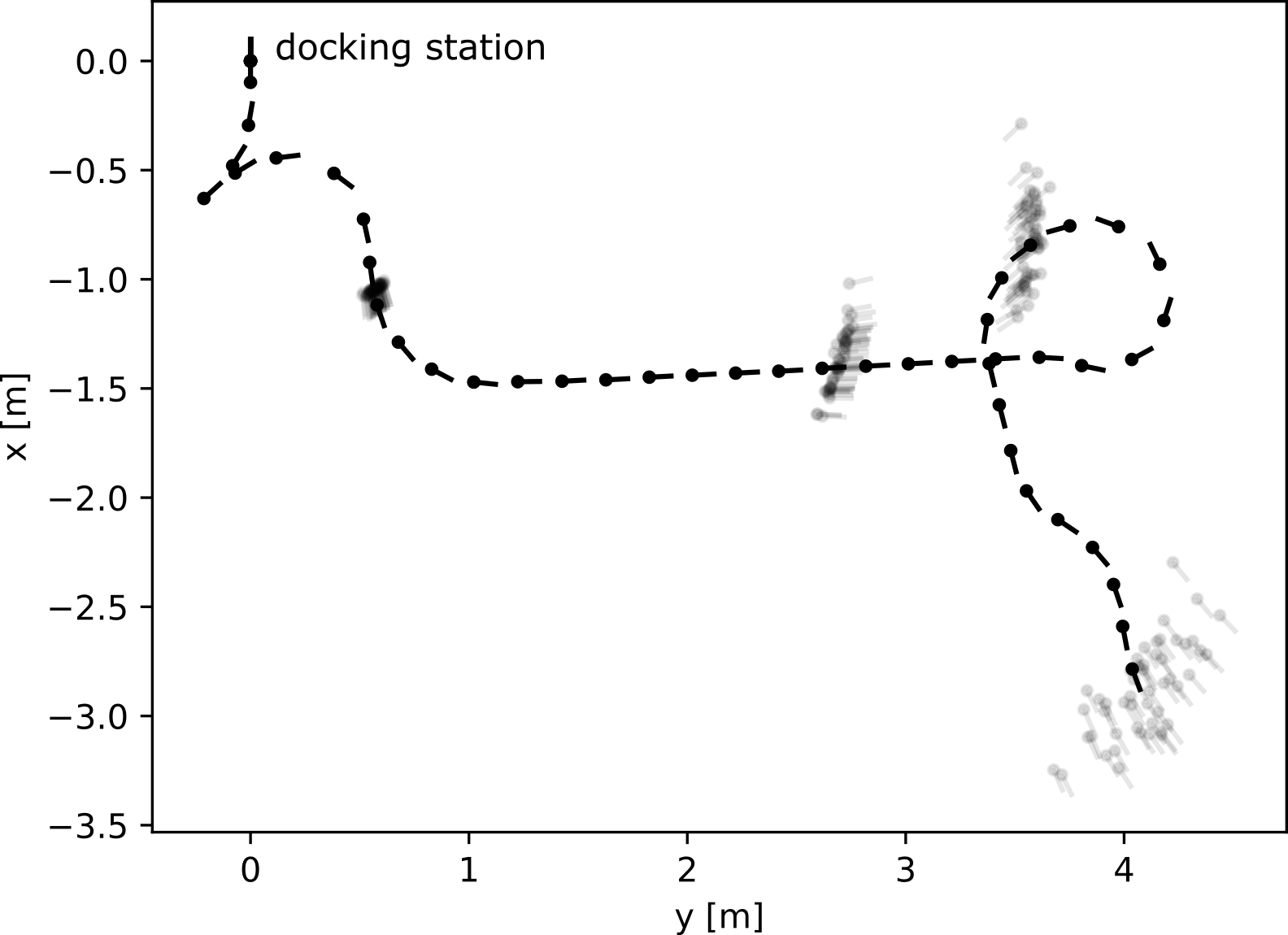}
    \caption{Visualization of measured inaccurate odometry (black), and 50 realizations of the uncertain odometry, 
    for the first minute of a training episode  of the RoboMaster robot.
    }
    \label{fig:rm_odometry}
\end{figure}

The robot onboard odometry module provides $\Tilde{\bm{p}}(t)$ w.r.t. the inertial frame ${\cal F}_0$, which coincides with ${\cal F}_r$ at the beginning of the acquisition. Figure~\ref{fig:rm_odometry} visualizes the inaccurate robot odometry as measured, and 50 realizations accounting for uncertainty, for the first minute of a training episode.

Furthermore, at time $t = 0$ the robot undocks from the docking station; similarly to the previous case, we use this information as a detector.
In this case study, the \gls{ooi} is the docking station and its frame ${\cal F}_i$ coincides with frame ${\cal F}_0$, as in Sec.~\ref{subsec:exp_setup_wall}.

We collect 20 episodes recording data at 15 Hz, for a total of 70k samples in 80 minutes (4 minutes per episode).
Each episode begins with the robot attached to the docking station; the robot is then teleoperated to explore the environment.
Data from different episodes are split into a training set (40k samples, 10 episodes), a validation set (15k samples, 5 episodes), and a testing set (15k samples, 5 episodes).
Training is performed with $\lambda_\text{o}=10$, similarly to Sec.~\ref{subsec:exp_setup_panda}.

\subsection{Model training}

In all case studies, a \gls{nn} is trained using Adam~\cite{adam} as optimizer with a learning rate of $1\text{e}^{-3}$; early stopping is used to determine when to conclude the training process.
An architecture based on MobileNet-V2~\cite{sandler2018mobilenet} with a total of 1 million parameters is used for the case studies in Sec.~\ref{subsec:exp_setup_panda} and Sec.~\ref{subsec:exp_setup_station};
it maps a $160 \times 120$ RGB image to an output vector representing a 3D pose (composed of 7 elements, 3 for the position and 4 for the quaternion).
In these two use cases, we artificially increase the amount of data used for the training, by adopting the following data augmentation techniques on the input images: blurring, multiplicative Gaussian noise, random brightness and contrast, random resized cropping.

In the case study detailed in Sec.~\ref{subsec:exp_setup_wall}, we use a simpler \gls{nn} architecture composed of four linear layers with a total of 1000 parameters, mapping the 7 infrared sensors' readings to a 3D pose.

\begin{figure}
    \subfloat[%
    \mdiff{Predictions (magenta-lime-cyan colored frame) and reconstructed ground-truth (red-green-blue frame) on testing set data.}]{
    \centering{
    \includegraphics[width=.31\linewidth]{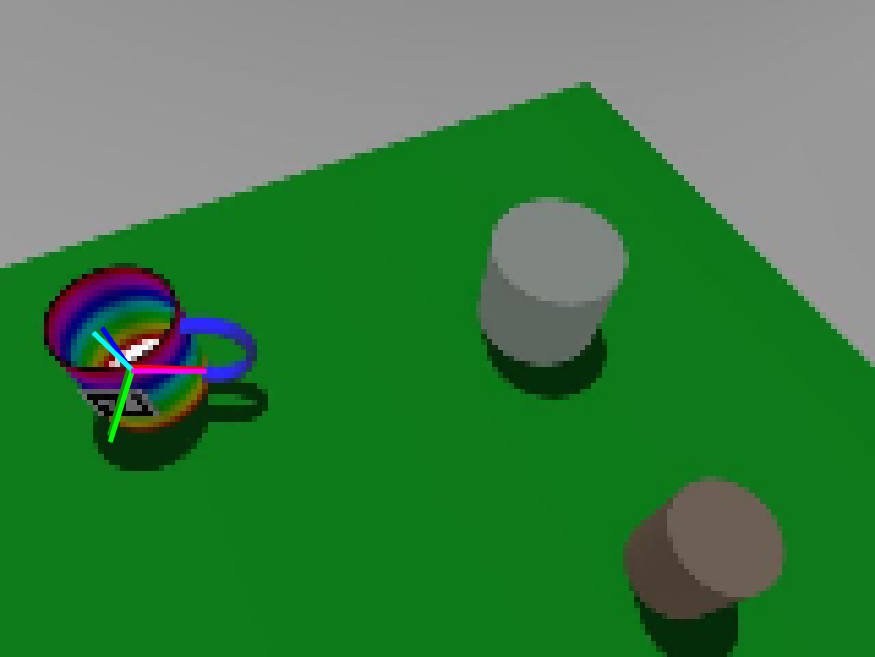}
    \hfill
    \includegraphics[width=.31\linewidth]{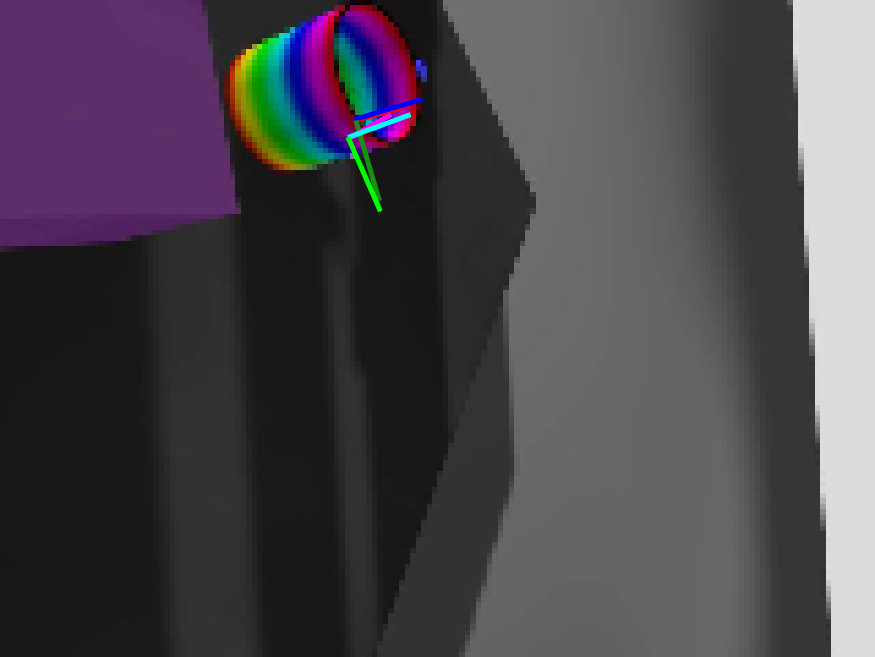}
    \hfill
    \includegraphics[width=.31\linewidth]{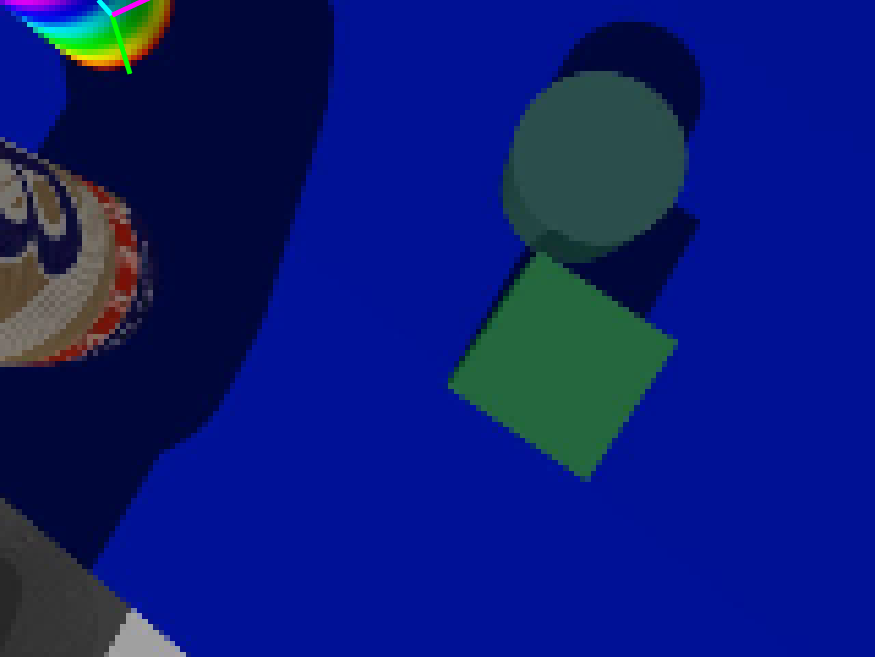}
    \label{fig:panda_pred}
    }
    }
    \\ [-7pt]
    \subfloat[%
    \mdiff{Input images from the additional testing scenario.}]{
    \centering{
    \includegraphics[width=.31\linewidth]{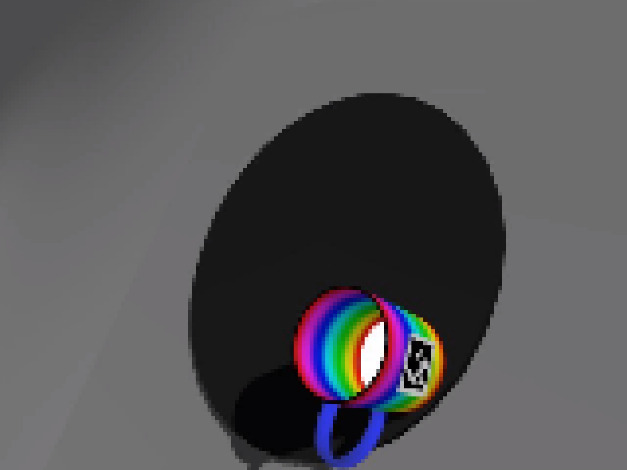}
    \hfill
    \includegraphics[width=.31\linewidth]{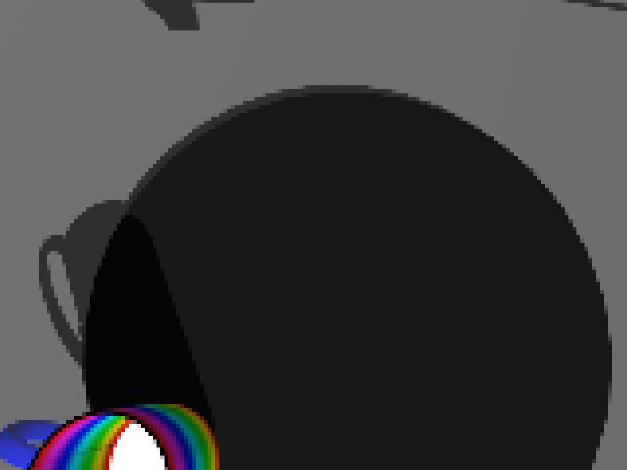}
    \hfill
    \includegraphics[width=.31\linewidth]{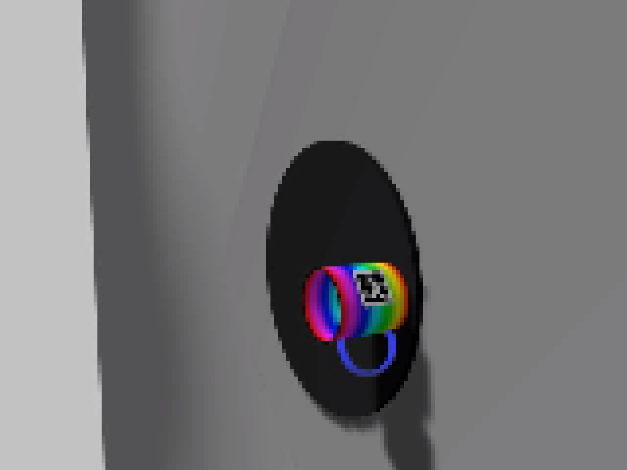}
    \label{fig:panda_pred_lp_snapshots}
    }
    }
    \\ 
    \raggedleft{
    \subfloat[\mdiff{Pose of the OOI in the inertial frame on the additional testing scenario.}]{
    \includegraphics[width=0.95\linewidth]{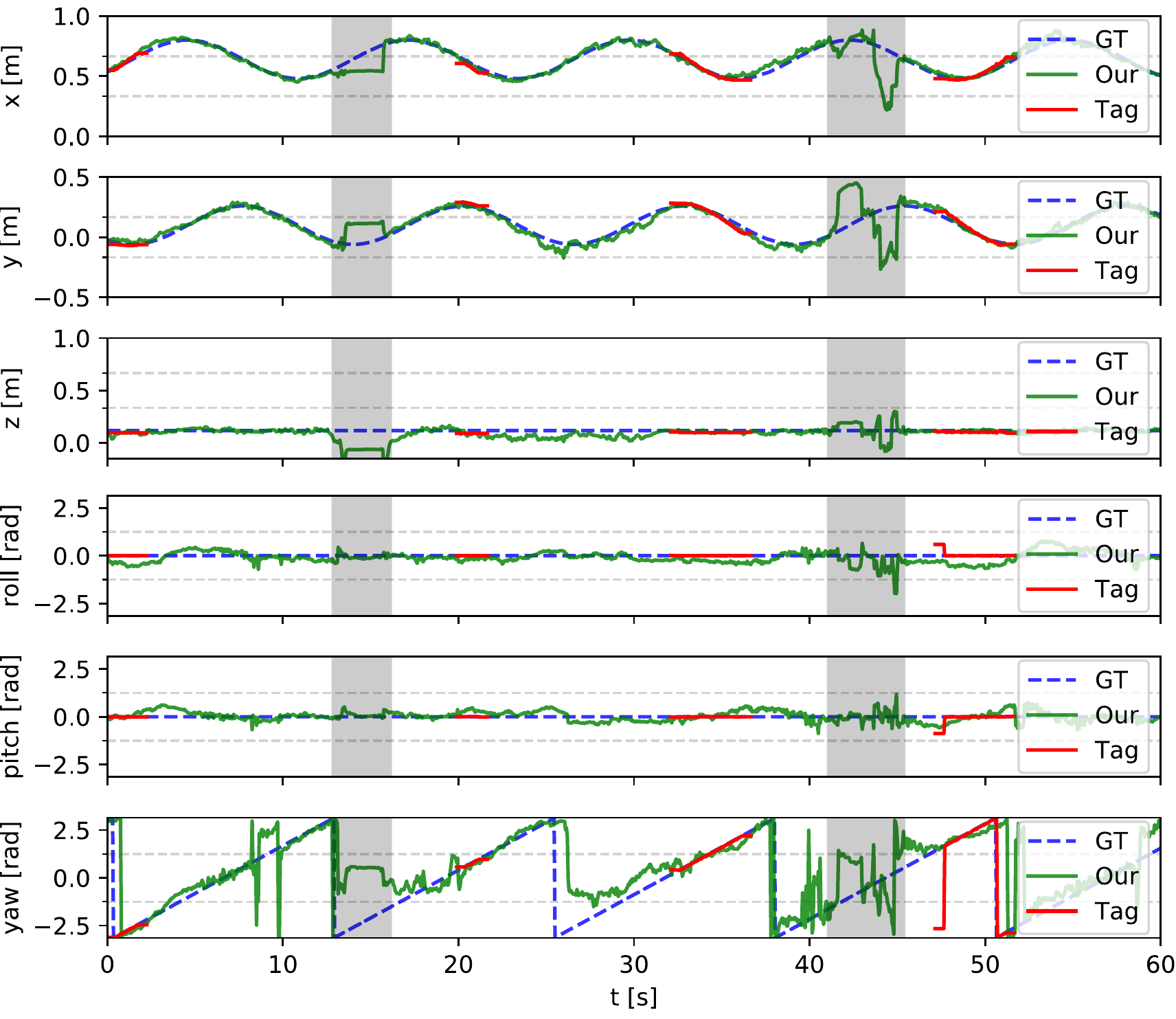}
    \label{fig:panda_pred_lp_ori}
    }
    }
    \caption{%
    \mdiff{Prediction of the \gls{ooi} pose with a robotic arm on the testing set (a); input images (b), ground-truth and predictions (c) relative to the additional testing scenario.
    In (b-c) a mug is placed on top of a rotating disk support.
    Grayed-out areas indicate time intervals in which the mug is not visible.
    }}
    \label{fig:panda_pred_lp}
\end{figure}


\section{Results and discussion}\label{sec:results}

\subsection{Object of Interest Pose Estimation with a Robotic Arm}\label{subsec:results_panda}

For the scenario described in Sec.~\ref{subsec:exp_setup_panda}, we first evaluate the trained model on the testing set, by comparing the predicted poses $\hat{\bm{y}}$ to the corresponding ground-truth $\bm{y}$.  Predictions occur independently for every frame; no information from state estimation or the detector is used in the process.

For the position component, the model achieves a Root Mean Squared Error of \SI{39.9}{\milli\meter}, and a coefficient of determination $R^2$ of $0.962$, $0.960$ and $0.866$ on the $x$, $y$ and $z$ components, respectively. The coefficient of determination is an adimensional measure of the quality of a regressor, which quantifies the amount of variance in the target variable that is explained by the model; an ideal regressor yields $R^2 = 1$, whereas a dummy regressor estimating the mean of the target variable yields $R^2 = 0$; we observe that, while all components are estimated well, the $z$ coordinate, i.e. the distance of the \gls{ooi} w.r.t. the camera, is estimated with lower accuracy; this is expected, since estimating distances is hard from low-resolution monocular images.
The rotational component is estimated with an average rotational error $\QUATDIST = 0.1417$, corresponding to an angle of about $\ang{25}$.

\mdiff{
As a comparison, we also trained a supervised model using
ground-truth poses of the \gls{ooi} instead of the detector outputs: the resulting prediction performance is the same as with the self-supervised approach. In fact, in the considered simulated environment the detector is very accurate and state estimation is ideal, which makes the self-supervised labels almost identical to ground-truth labels. This is not the case for the experiment in Section~\ref{subsec:results_wall}.}

Fig.~\ref{fig:panda_pred} shows a sequence of camera frames from the testing set with an overlay of the model's prediction and the ground-truth.
We observe that the model correctly identifies the object and accurately estimates its pose, even when the \gls{ooi} is only partially visible or occluded, and independently on the visibility of the fiducial marker used during training. 
When the object is not visible at all on the image, the model tends to confuse it with other objects that are visually similar; this highlights a limitation of our approach: we do not explicitly handle aliasing, i.e. the case in which $\bm{x}$ does not contain enough information to estimate $\bm{y}$.

To demonstrate the fact that the model, once trained, can be used in dynamic environments, we consider an additional testing scenario: the mug is placed on a rotating disk support and observed while the robot also moves. 
Fig.~\ref{fig:panda_pred_lp_snapshots} shows a sequence of camera frames captured during the experiment.
\mdiff{%
Fig.~\ref{fig:panda_pred_lp_ori} compares the model prediction to AprilTag detections, and to ground-truth.
For the sake of clarity, the target has been transformed in the frame ${\cal F}_0$,
along with the model predictions.}
The model (green in the plots) manages to predict a consistent pose (it well overlaps the blue dashed line of the ground-truth), outperforming the AprilTag detection (red line), which provides a measure for a small fraction of frames in which the tag is visible,
\mdiff{while our model runs at \SI{25}{Hz} on a GPU Nvidia Quadro P2000}.
The model produces good estimates of the mug pose from most points of view, even when it appears upside down.
Occasional failures occur when the mug is seen from a point of view that does not provide any reference to infer its actual rotation around the vertical axis -- 
i.e., when both the marker and the handle of the mug are not visible.
While it is robust to partial occlusions of the mug, the model fails to estimate the mug's pose when a very small portion of it is visible, or when it is totally invisible; time intervals in which the mug is not visible are depicted with a gray shadow in Fig.~\ref{fig:panda_pred_lp_ori}.

\subsection{Robot Heading Estimation using Infrared Sensors}\label{subsec:results_wall}
For the scenario described in Sec.~\ref{subsec:exp_setup_wall} we consider three possible sources of odometry: \emph{exact odometry}, where $\Tilde{\bm{p}}(t,u) = \bm{p}(t,u)$, as measured by the optitrack system, acting as an upper bound of the achievable performance; \emph{pointwise odometry}, where the relative pose $\Tilde{\bm{p}}(t,u)$ is computed according to the known robot kinematics and readings of the wheel rotation sensors between timesteps $t$ and $u$; \emph{uncertain odometry}, where the probability distribution $\Tilde{\bm{P}}(t,u)$ is approximated by \mdiff{$N_{\text{mc}}$ = 50 realizations} of the odometry between timesteps $t$ and $u$, obtained by corrupting readings of the wheel rotation sensors with white Gaussian noise, whose variance matches the known uncertainty of the sensor.

\begin{figure}
    \centering
    \includegraphics[width=0.91\linewidth]{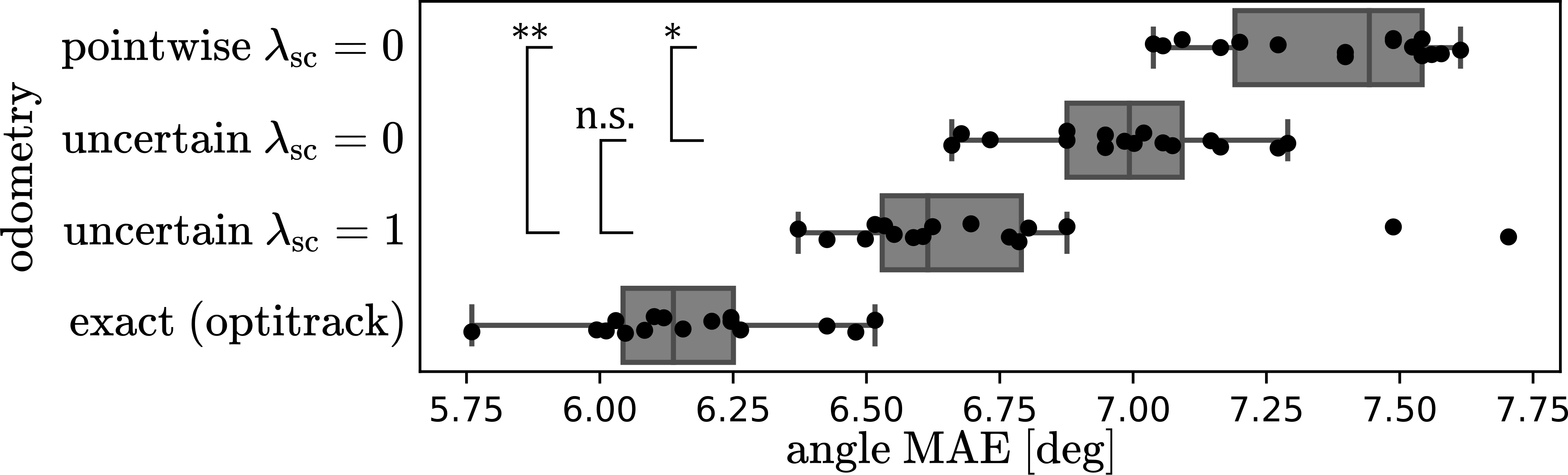}
    \caption{Mean absolute angle error (lower is better) in the heading estimation task. $^*\;p = 0.0003$; $^{**}\;p = 0.0008$; n.s. means not significant ($p = 0.029$).}
    \label{fig:thymio_box}
\end{figure}

Fig.~\ref{fig:thymio_box} reports the Mean Absolute Error between the predicted angle of the robot pose with respect to the wall, against the ground-truth angle as measured by the optitrack system; this metric is reported for four models trained with different odometry sources, with ($\lambda_\text{sc} = 1$) or without ($\lambda_\text{sc} = 0$) the state-consistency loss.  Each of the four models is trained and evaluated 16 times according to the leave-one-episode-out cross-validation scheme.  We observe that: i) using uncertain odometry instead of pointwise odometry improves the prediction performance of the resulting model;\footnote{We use the non-parametric Wilcoxon signed-rank test between matched samples, i.e. the performance of two models on the same cross-validation fold ($p = 0.0003$).} ii) additionally enforcing the state-consistency loss ($\lambda_\text{sc} = 1$) further improves performance, even though the incremental improvement over the uncertain model with $\lambda_\text{sc} = 0$ is not statistically significant; iii) compared to the model trained with exact odometry,
\mdiff{serving as a supervised learning upperbound}%
, the performance gap of our best model is less than half of the performance gap of the baseline model (pointwise $\lambda_\text{sc} = 0$), i.e. \ang{0.57} vs \ang{1.20}.

\subsection{Indoor Localization of a Ground Robot}\label{subsec:results_station}
For the scenario described in Sec.~\ref{subsec:exp_setup_station}, we report one qualitative and one quantitative experiment.

\begin{figure}
    \centering
    \includegraphics[width=0.84\linewidth]{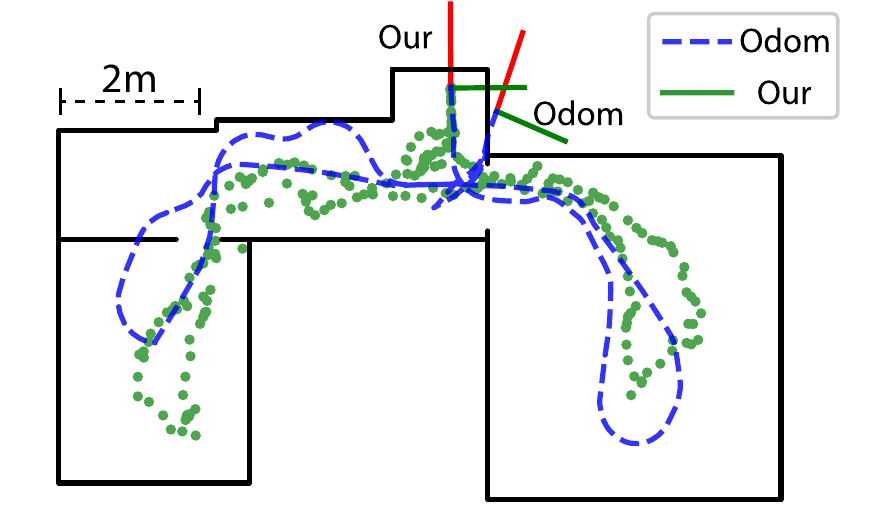}
    \caption{Robot localization task on testing data.}
    \label{fig:station_pred}
\end{figure}

In the qualitative experiment, we train a model using uncertain odometry \mdiff{($N_{\text{mc}} = 50$)} and $\lambda_\text{sc} = 1$, then apply the trained model independently on each frame of a testing episode.
Fig.~\ref{fig:station_pred} shows the position component of the predicted poses (green), compared to the poses returned by the robot's odometry (blue) for the same trajectory; despite the short duration of the testing episode (about 4 minutes), the drift of the odometry trajectory is apparent (e.g. the blue trajectory passes through a wall on the left); in comparison, the poses predicted by our approach are locally noisy but not affected by accumulating error.  The figure also depicts the predicted pose at the end of the trajectory for each of the two methods; our model correctly predicts that the robot is back at the docking station, whereas the odometry has drifted by about \SI{0.5}{\meter} and \ang{20}.

In the quantitative experiment, we consider four timesteps $t \in {t_1, t_2, t_3, t_4}$ of the testing episode, whose ground-truth poses $\bm{p}(t)$ have been manually measured with respect to the docking station.  For each of the four timesteps, we measure the positional and rotation components of the error against such ground-truth, for: i) the pose $\Tilde{\bm{p}}(t)$ estimated by the robot odometry; ii) the poses estimated by each of four models.  In particular, we consider models trained with pointwise or uncertain odometry, with or without using the state-consistency loss. Table~\ref{tab:scenarios} shows that: using the state-consistency loss improves both positional and rotational errors; using uncertain odometry during training consistently outperforms pointwise odometry.

\begin{table}[htbp]
    \centering
    \caption{Quantitative performance on the indoor localization task}
    \begin{tabular}{lrr}
    \toprule
    Method & Position Error [mm] & Rotation Error [deg] \\
    \midrule
    Odometry & 84.9 & 11.9 \\
    Pointwise, $\lambda_\text{sc} = 0$ & 62.8 & 23.8 \\
    Pointwise, $\lambda_\text{sc} = 1$ &  49.1 & 7.5 \\
    Uncertain, $\lambda_\text{sc} = 0$ & 73.5 & 9.1 \\
    Uncertain, $\lambda_\text{sc} = 1$ & \textbf{35.2} & \textbf{3.8} \\
    \bottomrule
    \end{tabular}
    \label{tab:scenarios}
\end{table}

\section{Conclusions}\label{sec:conclusions}
We presented a general \mdiff{self-supervised learning approach for spatial perception tasks}, and instantiated it on three case studies.
The approach is general enough to be applicable to different robots and sensor apparatus, requiring only a possibly uncertain odometry and a detector that sparsely produces a ground-truth estimate.
A novel loss allows us to evaluate the model outputs
also for timesteps in which no supervision is available, by propagating such supervision from different timesteps using uncertain state estimates.
Furthermore, the loss formulation enforces consistency among predictions at different timesteps, which further improves performance.
Results show consistent and statistically significant improvements of models learned with the uncertainty-aware version of the loss compared to the baseline.

\bstctlcite{IEEEexample:BSTcontrol}

\bibliographystyle{IEEEtran}
\bibliography{references}

\end{document}